\documentclass[sigconf,nonacm]{acmart}

\renewcommand\footnotetextcopyrightpermission[1]{} 

\usepackage{amsmath}
\usepackage{graphicx}
\usepackage{xcolor}
\usepackage{subfigure}
\usepackage{diagbox}
\usepackage{multirow}
\usepackage{algorithm}
\usepackage{listings}

\usepackage{amssymb}

\newenvironment{myequation}{
\begin{equation}
}{
\end{equation}
}

\AtBeginDocument{%
  \providecommand\BibTeX{{%
    \normalfont B\kern-0.5em{\scshape i\kern-0.25em b}\kern-0.8em\TeX}}}

\setcopyright{acmcopyright}
\copyrightyear{2018}
\acmYear{2018}
\acmDOI{10.1145/1122445.1122456}

\acmConference[WWW '22]{}{April 2022}{Lyon, France}
\acmPrice{15.00}
\acmISBN{978-1-4503-XXXX-X/18/06}

\begin{document}

\title{Contrastive Adaptive Propagation Graph Neural Networks for Efficient Graph Learning}


\author{Jun Hu$^{1}$, Shengsheng Qian$^{1,2}$, Quan Fang$^{1,2}$, Changsheng Xu$^{1,2,3}$}
\affiliation{
  \institution{$^{1}$National Laboratory of Pattern Recognition, Institute of Automation, Chinese Academy of Sciences}
  \country{}
}
\affiliation{
  \institution{$^{2}$School of Artificial Intelligence, University of Chinese Academy of Sciences}
  \country{}
}
\affiliation{
  \institution{$^{3}$Peng Cheng Laboratory, ShenZhen, China}
  \country{}
}
\email{hujunxianligong@gmail.com, {shengsheng.qian, qfang, csxu}@nlpr.ia.ac.cn}

\renewcommand{\shortauthors}{Jun Hu, Shengsheng Qian, Quan Fang, Changsheng Xu}

\begin{abstract}
Graph Neural Networks (GNNs) have achieved great success in processing graph data by extracting and propagating structure-aware features.
Existing GNN research designs various propagation schemes to guide the aggregation of neighbor information.
Recently the field has advanced from local propagation schemes that focus on local neighbors towards extended propagation schemes that can directly deal with extended neighbors consisting of both local and high-order neighbors.
Despite the impressive performance, existing approaches are still insufficient to build an efficient and learnable extended propagation scheme that can adaptively adjust the influence of local and high-order neighbors.
This paper proposes an efficient yet effective end-to-end framework, namely Contrastive Adaptive Propagation Graph Neural Networks (CAPGNN), to address these issues by combining Personalized PageRank and attention techniques.
CAPGNN models the learnable extended propagation scheme with a polynomial of a sparse local affinity matrix, where the polynomial relies on Personalized PageRank to provide superior initial coefficients. 
In order to adaptively adjust the influence of both local and high-order neighbors,  a coefficient-attention model is introduced to learn to adjust the coefficients of the polynomial.
In addition, we leverage self-supervised learning techniques and design a negative-free entropy-aware contrastive loss to explicitly take advantage of unlabeled data for training. 
We implement CAPGNN as two different versions named CAPGCN and CAPGAT, which use static and dynamic sparse local affinity matrices, respectively.
Experiments on graph benchmark datasets suggest that CAPGNN can consistently outperform or match state-of-the-art baselines.
The source code is publicly available at https://github.com/hujunxianligong/CAPGNN.

\end{abstract}

\ccsdesc[300]{Information systems~Data mining}

\keywords{Graph Neural Networks, Graph Deep Learning, Graph Contrastive Learning}


\maketitle

\section{Introduction}\label{sec:introduction}

Graphs are powerful and expressive data structures, and they serve as a common language for modeling relational data, such as knowledge and social networks.
In recent years, Graph Neural Networks (GNNs) have been proposed for deep learning on graph data.
Due to GNNs' demonstrably powerful ability, they have attracted researchers and developers from a wide range of domains, and have been successfully applied to applications such as recommendation systems~\cite{DBLP:conf/www/Fan0LHZTY19,DBLP:conf/www/WangZLLZLZ20,DBLP:conf/www/TanLZYZH20,DBLP:conf/www/ZhengGCJL21,cai2021grecx}, community question answering~\cite{DBLP:conf/mm/HuQFX19,DBLP:conf/aaai/0056CDWZ021}, social network analysis~\cite{DBLP:conf/www/PiaoZXCL21,DBLP:conf/www/SankarLYS21}, and social media understanding~\cite{DBLP:conf/mir/WangQHFX20,DBLP:conf/mm/WeiWN0HC19}.

GNNs can aggregate neighbor information to enrich the semantics of vertices for better vertex representations.
Existing GNN research designs different propagation schemes to guide the aggregation of neighbor information.
Recently the field has advanced from local propagation schemes that focus on local neighbors towards extended propagation schemes that can directly deal with extended neighbors consisting of both local and high-order neighbors~\cite{DBLP:conf/nips/KlicperaWG19}.
The former research such as GNNs such as GCN~\cite{DBLP:conf/iclr/KipfW17} and GAT~\cite{DBLP:conf/iclr/VelickovicCCRLB18} design GNN layers to learn local propagation schemes, and they perform high-order propagation with naive deep architectures, which simply stack multiple GNN layers together and usually suffer from the over-smoothing problem, where the learned representation of vertices may be indistinguishable.
The later research such as APPNP~\cite{DBLP:conf/iclr/KlicperaBG19} and AGCN~\cite{DBLP:conf/aaai/LiWZH18} can explicitly compute the influence of both local and high-order neighbors for propagation, and it can alleviate the over-smoothing problem with techniques such as Personalized PageRank~\cite{Page1999ThePC} and structure learning~\cite{DBLP:journals/corr/abs-2103-03036}.

\begin{figure}[t]
\centering\includegraphics[scale=0.41]{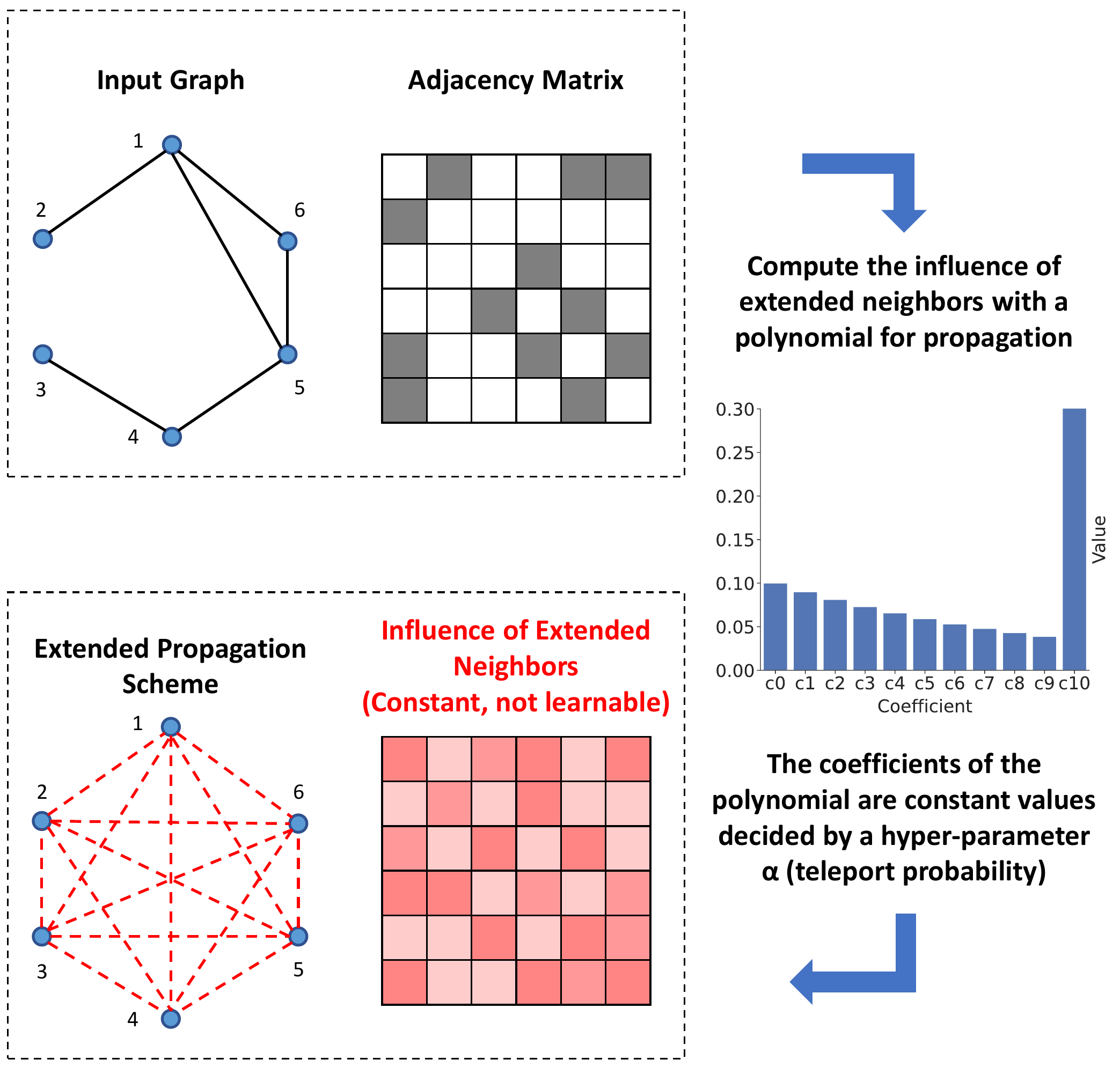}
\caption{An example of APPNP's extended propagation scheme.}
 \label{fig:intro_coefs}
\end{figure}

Although many recent GNNs can achieve impressive performance with extended propagation schemes, they usually suffer from two limitations and still have room for improvement.
(1) First, existing approaches are insufficient to build an efficient learnable extended propagation scheme that can adaptively adjust the influence of local and high-order neighbors.
Although approaches such as APPNP~\cite{DBLP:conf/iclr/KlicperaBG19} and GRAND~\cite{DBLP:conf/nips/FengZDHLXYK020} can build efficient extended propagation scheme by leveraging Personalized PageRank or RandomWalk techniques, most of them lack of learnable components that can adaptively adjust the influence of high-order neighbors.
For example, as shown in Figure~\ref{fig:intro_coefs}, APPNP computes the influence of extended neighbors for propagation with a polynomial that only relies on the adjacency matrix and a teleport probability $\alpha$, where the former is constant input data and the later is a unlearnable hyper-parameter that decides the coefficients of the polynomial.
As a result, the model iteself is not able to learn to adaptively adjust the influence of extended neighbors.
%
%
Different from these approaches, another research line, such as AGCN~\cite{DBLP:conf/aaai/LiWZH18} and GAUG-M~\cite{DBLP:conf/aaai/0003LNW0S21}, highly relies on learnable structure learning components to learn implicit relations between extended neighbors, which are used as the influence of extended neighbors for the propagation schemes.
However, directly applying structure learning on extended neighbors may result in high time and space complexity.
For example, in order to learn the influence of extended neighbors, approaches such as AGCN~\cite{DBLP:conf/aaai/LiWZH18} and GAUG-M~\cite{DBLP:conf/aaai/0003LNW0S21} compute the similarity scores between every vertex pair and construct edges between vertex pairs with high similarity scores, which is computational expensive even on small datasets.
(2) A great number of real-world graph data only provide limited label information, and the standard semi-supervised training strategy~\cite{DBLP:conf/iclr/KipfW17} of GNNs should be improved to support the training of extended propagation schemes.
One of the reasons is that learning extended propagation schemes may involve more complicated model architectures, which may require more supervised information for optimization.
Under the standard semi-supervised training strategy, the supervised loss function is only explicitly applied on labeled data, and the unlabeled data contribute only implicitly for the optimization.
Therefore, it is necessary to leverage techniques such as self-supervised learning to design explicit loss functions for unlabeled data, which is likely to allow us to learn extended propagation schemes with limited label information.

To address these limitations, we propose an efficient yet effective end-to-end framework, namely Contrastive Adaptive Propagation Graph Neural Networks (CAPGNN), which can obtain learnable extended propagation schemes that can adaptively adjust the influence of local and high-order neighbors with both labeled and unlabeled data.
CAPGNN models the learnable extended propagation scheme with a polynomial of a sparse local affinity matrix, where the polynomial relies on Personalized PageRank to provide superior initial coefficients. 
In order to adaptively adjust the influence of local and high-order neighbors, we design a coefficient-attention model to learn to adjust the coefficients of the polynomial.
To retain the sparsity of the graph for efficient computation, we decompose the polynomial and implement the extended propagation as efficient power iterations of a sparse matrix.
To better optimize the extended propagation scheme with limited labeled information, we propose a negative-free entropy-aware contrastive loss to explicitly take advantage of unlabeled data for training.
The proposed contrastive loss does not require negative samples, and it simultaneously takes into account the semantic consistency property of vertex representations and a low-entropy assumption for semi-supervised learning.

In short, we conclude our main contributions as follows:
\begin{itemize}
\item We propose an efficient yet effective end-to-end framework, namely Contrastive Adaptive Propagation Graph Neural Networks (CAPGNN), to obtain learnable extended propagation schemes that can adaptively adjust the influence of local and high-order neighbors with both labeled and unlabeled data.
%
%
\item Our framework models the learnable extended propagation scheme with a polynomial of a sparse affinity matrix, which can leverage Personalized PageRank to provide superior initial coefficients and can adaptively adjust the influence of high-order neighbors with a coefficient-attention model.
%
%
\item We propose a negative-free entropy-aware contrastive loss to explicitly take advantage of unlabeled data for better optimization of the extended propagation scheme.
The loss allows our model to perform contrastive learning without negative samples, and it simultaneously takes into account the semantic consistency property of vertex representations and a low-entropy assumption for semi-supervised learning.
%
\item We conduct experiments on benchmark datasets, and the experimental results show that our models consistently outperform or match state-of-the-art baselines.
\end{itemize}

\section{Related Work}\label{sec:related_work}

Graph Neural Networks can take advantage of both the content (vertex features) and the structure (edge information) of graphs for graph deep learning.
Kipf et al. develop an end-to-end model named Graph Convolution Networks (GCN)~\cite{DBLP:conf/iclr/KipfW17}, which can project graph data into the spectral domain with graph Fourier transform and then perform convolutional operations in the spectral domain.
Each GCN layer is implemented in the form of message propagation between local neighbors and it relies on the full structure of graphs for computation; therefore, it is sometimes difficult to extend to large graphs due to large computational complexity.
FastGCN~\cite{DBLP:conf/iclr/ChenMX18}, GraphSage~\cite{DBLP:conf/nips/HamiltonYL17}, and AS-GCN~\cite{DBLP:conf/nips/Huang0RH18} propose to utilize sampling strategies on graphs to alleviate the scalability problem of GNNs.
%
%
Graph Attention Network (GAT)~\cite{DBLP:conf/iclr/VelickovicCCRLB18} and Attention-based Graph Neural Network (AGNN)~\cite{thekumparampil2018attentionbased} introduce attention techniques into GNNs to dynamically learn attention weights among neighbors as local propagation schemes.
%
%
The aforementioned approaches focus on learning local propagation schemes, and their GNN layers are designed to perform message propagation between local neighbors.
In order to exploit high-order neighbor information within $K$ hops, we need to stack $K$ GNN layers and add non-linearity to establish a naive deep architecture.
However, research shows that such deep architecture may suffer from the over-smoothing problem, where the learned representation of vertices may become indistinguishable as the architecture becomes deeper~\cite{DBLP:journals/corr/abs-2008-09864}.
As a result, the architecture is not able to effectively exploit high-order information and it usually exhibits degraded performance when $K > 2$.
Although the trick that adding residual connections between hidden GNN layers can effectively facilitate the training of the deep architecture, it merely slightly alleviates the over-smoothing problem by slowing down the over-smoothing process~\cite{DBLP:conf/iclr/KipfW17,DBLP:conf/iclr/ZhuK21}.
Instead of simply stacking multiple GNN layers, a lot of recent GNN models explore other strategies that can effectively take advantage of high-order neighbor information.

Recently, GNN research has advanced from the above local propagation schemes towards extended propagation schemes that can explicitly model the influence of extended neighbors consisting of both local and high-order neighbors for propagation~\cite{DBLP:conf/nips/KlicperaWG19}.
%
%
%
With the guidance of carefully designed extended propagation schemes, GNNs are able to alleviate the over-smoothing problem and effectively exploit high-order neighbor information.
Abu-El-Haija et al. propose MixHop~\cite{DBLP:conf/icml/Abu-El-HaijaPKA19} to aggregate features of extended neighbors at various distances with different powers of the adjacency matrix in each GNN layer, and then mix them with the concatenation operation.
%
%
%
Klicpera et al. propose Personalized Propagation of Neural Predictions (PPNP)~\cite{DBLP:conf/iclr/KlicperaBG19} to model the extended propagation scheme with a Personalized PageRank matrix, which is able to preserve the personalized vertex information while aggregating extended neighbor information.
They also propose a fast approximation of PPNP named Approximate Personalized Propagation of Neural Predictions (APPNP)~\cite{DBLP:conf/iclr/KlicperaBG19}, which implements the Personalized PageRank-based propagation via efficient power iterations.
Wu et al. propose the Simple Graph Convolution (SGC)~\cite{DBLP:conf/icml/WuSZFYW19} to simplify multiple GCN layers into a single GNN layer with the power of local propagation matrix, which enables SGC to effectively aggregate high-order neighbors.
Zhu et al. modify Markov Diffusion Kernel to derive a variant of GCN called Simple Spectral Graph Convolution (S$^2$GC), which combines different powers of local propagation matrices and can trade off the global and local contexts of vertices.
Lim et al. propose Class-Attentive Diffusion Network (CAD-Net)~\cite{DBLP:conf/aaai/LimUCJC21} to trade-off between vertices' own features and the features aggregated from extended neighbors depending on the local class-context.
%
%
Different with these approaches, another research line focuses on learning to extend neighbors with structure learning techniques~\cite{DBLP:journals/corr/abs-2103-03036}, which can dynamically learn edges between arbitraty vertex pairs based on features such as vertex representations.
%
%
AGCN~\cite{DBLP:conf/aaai/LiWZH18} and GAUG-M~\cite{DBLP:conf/aaai/0003LNW0S21} propose to use learnable edge predictors to capture implicit edges between high-order neighbors for neighbor extension, which allow these models to propagate information between high-order neighbors with shallow networks.
ADSF-RWR~\cite{DBLP:conf/iclr/0001ZWZ20} designs an attention model to take advantage of both structural and content information to extend neighbors for each vertex from a receptive field composed of the vertex's high-order neighbors within a specified distance.

In addition to extended propagation schemes, some recent research also explores ways to exploit unlabeled data for better training of GNNs.
We are also interested in these approaches since high-order propagation schemes may invoke more complex architectures and they may rely on sufficient data for training.
Inspired by Deep Infomax (DIM)~\cite{DBLP:conf/iclr/HjelmFLGBTB19}, Velickovic et al. propose an unsupervised manner named Deep Graph Infomax (DGI) ~\cite{DBLP:conf/iclr/VelickovicFHLBH19}, which can learn vertex representations through local-global (vertex-graph) mutual information maximization.
Hassani et al. introduce a self-supervised approach~\cite{DBLP:conf/icml/HassaniA20} for vertex and graph level representation learning by contrasting different structural views of a graph.
Zhu et al. design deep GRAph Contrastive rEpresentation
learning (GRACE)~\cite{DBLP:journals/corr/abs-2006-04131}, which generates two views of graphs via corruption and learns vertex representations through maximizing the agreement of vertex representations in these two views.
These approaches are designed for unsupervised representation learning.
Recently, some approaches propose to apply regularizations on unlabeled data for better semi-supervised learning.
GraphMix~\cite{DBLP:conf/aaai/VermaQKLBKT21} proposes to use an interpolation function to create new unlabeled data and apply regularizations on them.
GRAND~\cite{DBLP:conf/nips/FengZDHLXYK020} leverages consistency regularization to optimize the consistency of vertex representations across different augmented views.

Different from existing approaches, our framework models the learnable extended propagation scheme with a polynomial of a sparse local affinity matrix, can adaptively adjust the influence of local and high-order neighbors with a coefficient-attention model.
Moreover, we propose a negative-free entropy-aware contrastive loss to explicitly take advantage of unlabeled data for training, which can simultaneously take into account the semantic consistency property of vertex representations and a low-entropy assumption for semi-supervised learning.

\begin{figure*}[t]
  
  \centering\includegraphics[scale=0.45]{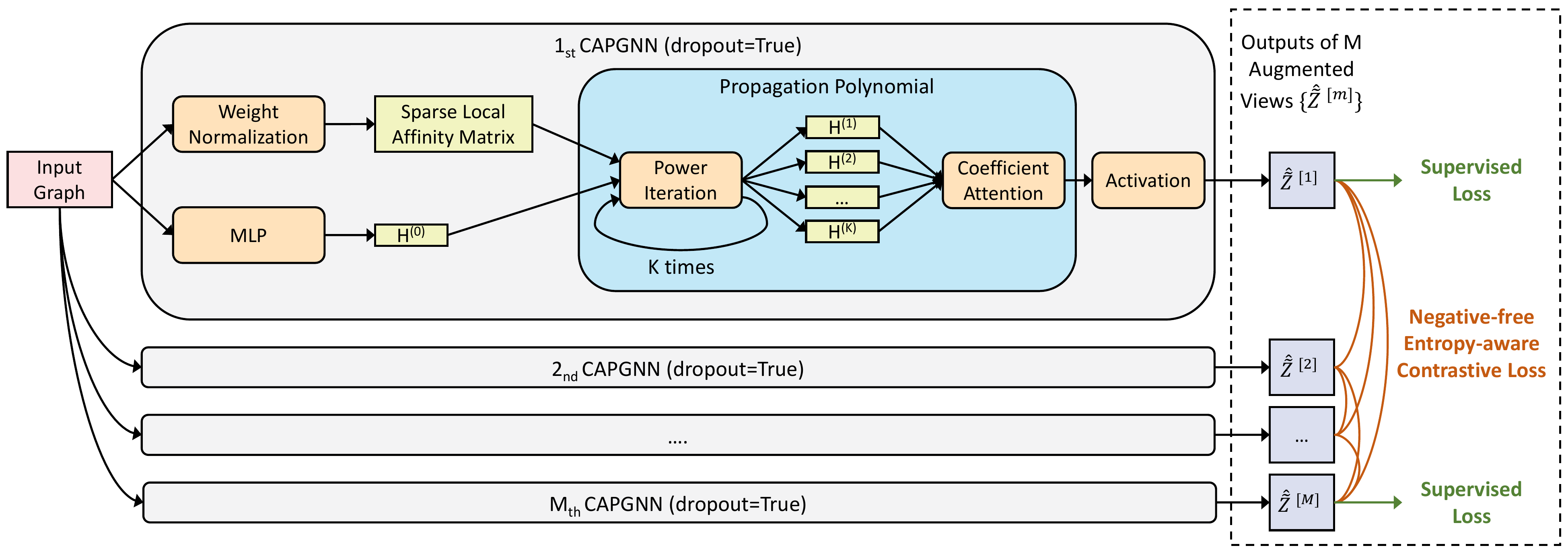}
  \caption{Overall Framework of CAPGNN.}
   \label{fig:framework}
  \end{figure*}

\section{Method}

\subsection{Problem Definition}


We use $\mathcal{G} = ( V, E )$ to denote a graph, where $V$ and $E$ are the sets of vertices and edges, respectively.
We use $|V|$ and $|E|$ to denote the number of vertices and edges, respectively.
For computation, we use another form $G=(X, A)$ to represent the graph, where $X \in \mathbb{R}^{|V| \times d_{x}}$ is the feature matrix of vertices and $A \in \mathbb{R}^{|V| \times |V|}$ is the adjacency matrix.
The $i_{th}$ row of $X$ is the $d_{x}$-dimensional feature vector of the $i_{th}$ vertex $v_i$, and each element $A_{ij}$ of $A$ is the weight of the edge from vertex $v_i$ to $v_j$.
Given a graph $G$, our model aims to predict the labels of test vertices based on $X$ and $A$.

\subsection{Overall Framework}

Figure ~\ref{fig:framework} shows the overall framework of our model, which mainly consists of the following two components:
\begin{itemize}
\item \textbf{Adaptive Propagation Graph Neural Networks:} As shown in the first grey box in the left part of Figure ~\ref{fig:framework}, we build a network to model the learnable extended propagation scheme with a polynomial of a sparse affinity matrix, which can leverage Personalized PageRank to provide superior initial coefficients and can adaptively adjust the influence of high-order neighbors with a coefficient-attention model.
\item \textbf{Negative-free Entropy-aware Graph Contrastive Loss:} As shown in the right part of Figure ~\ref{fig:framework}, we design a negative-free entropy-aware contrastive loss to explicitly take advantage of unlabeled data for better optimization, which simultaneously takes into account the semantic consistency property of vertex representations and a low-entropy assumption for semi-supervised learning.
\end{itemize}

\subsection{Adaptive Propagation Graph Neural Networks}

Some network representation learning theories~\cite{DBLP:conf/www/TangQWZYM15} assume that vertices that have similar context distribution are likely to share similar semantics, where the context refers to the neighbors of vertices. 
Inspired by this, our model learns an extended propagation scheme to capture the distribution of vertices' extended neighbors, and then uses the propagation scheme to guide the aggregation of neighbors for vertex representation learning.
In order to efficiently learn a propagation scheme for extended neighbors, our model first captures the local context distribution with a sparse local affinity matrix $\hat{A}$, and then extends it for extended neighbors with a polynomial of $\hat{A}$.


\subsubsection{Sparse Local Affinity Matrix.}
CAPGNN includes two implementations named \textbf{CAPGCN} and \textbf{CAPGAT}, which use two different types of sparse local affinity matrix $\hat{A}_{gcn}$ and $\hat{A}_{gat}$ to capture the local context distribution of vertices, respectively.
%
$\hat{A}_{gcn}$ is a static normalized local affinity matrix used by GCN~\cite{DBLP:conf/iclr/KipfW17}, which is computed as follows:
\begin{myequation}
  \hat{A}_{gcn} = \tilde{D}^{-\frac{1}{2}}(A + I)\tilde{D}^{-\frac{1}{2}}
\end{myequation}
where $\tilde{D}_{ii} = \sum_{j} A_{ij} + 1$ is the degree matrix of $A + I$.
$\hat{A}_{gat}$ is a dynamic local affinity matrix that combines $\hat{A}_{gcn}$ with the graph attention technique, which is computed as follows:
%
%
\begin{myequation}
\begin{split}
\hat{A}_{gat} & = \beta \hat{A}_{gcn} + (1 - \beta) \tilde{D}^{\frac{1}{2}} \Upsilon \tilde{D}^{-\frac{1}{2}} \\
& = \beta \tilde{D}^{-\frac{1}{2}}(A + I)\tilde{D}^{-\frac{1}{2}} + (1 - \beta) \tilde{D}^{\frac{1}{2}} \Upsilon \tilde{D}^{-\frac{1}{2}} 
\end{split}
\end{myequation}
where $\Upsilon$ is the local affinity matrix obtained with GAT~\cite{DBLP:conf/iclr/VelickovicCCRLB18} and $\beta \in \mathbb{R}$ is hyper-parameter between 0 and 1 that can trade off static and dynamic information.
$\Upsilon$ is normalized by GAT with a row-level normalization operation, and it can be considered as an approximation of $\tilde{D}^{-1}(A + I)$ rather than $\tilde{D}^{-\frac{1}{2}}(A + I)\tilde{D}^{-\frac{1}{2}}$.
Therefore, it is improper to directly combine $\Upsilon$ and $\hat{A}_{gcn}$.
To address this problem, we design a renormalization trick and introduce $\tilde{D}^{\frac{1}{2}} \Upsilon \tilde{D}^{-\frac{1}{2}}$, which shares the same normalization form of $\hat{A}_{gcn}$.

\subsubsection{Adaptive Extended Propagation Scheme.}\label{sec:detail_adapt_prop}
%
We model the adaptive extended propagation scheme in the form of a polynomial $\hat{\tilde{A}}' = \sum_{k=0}^{K} c_k (\hat{A}')^{k}$, where $\hat{A}'$ denotes the transpose of $\hat{A}$, ${c_k}$ are the coefficients of the polynomial, and the element $\hat{\tilde{A}}'_{ij}$ of $\hat{\tilde{A}}'$ represents the propagation weight from vertex $v_i$ to vertex $v_j$.
Research~\cite{DBLP:conf/iclr/KlicperaBG19} shows that Personalized PageRank can provide superior prior extended propagation schemes for GNNs.
Therefore, instead of directly optimizing ${c_k}$, our model initializes ${c_k}$ based on Personalized PageRank and learns to adaptively adjust ${c_k}$ with a coefficient-attention model.

We first construct $K + 1$ polynomials $\{U^{(k)} | 0 \leq k \leq K\}$ based on Personalized PageRank as follows:
\begin{myequation}
  U^{(0)} = I
\end{myequation}
%
%
\begin{myequation}
\begin{split}
  U^{(k)} & = (1 - \alpha) \hat{A}' U^{(k-1)} + \alpha I \\
          & = (1 - \alpha)^k (\hat{A}')^k +  \sum_{l=1}^{k} \alpha (1 - \alpha)^{l-1} (\hat{A}')^{l-1}
\end{split}
\end{myequation}
where $U^{(k)}$ corresponds to the $k-hop$ propagation scheme and the element $U^{(k)}_{ij}$ represent the propagation weight from vertex $v_i$ to vertex $v_j$.
Note that for each $U^{(k)}$, the coefficients $\{ c_l | 0 \leq l \leq k \}$ are constant values determined by the hyper-parameter $\alpha$, which are not learnable.

In order to build a robust and learnable extended propagation scheme $\hat{\tilde{A}}'$, our model combines the coefficients of $\{U^{(k)} | 1 \leq k \leq K \}$ and introduces attention-based learnable coefficients $\{s_{k} | 1 \leq k \leq K \}$ to adjust the combined coefficients as follows:
\begin{myequation}\label{eq:combined_A}
\begin{split}
\hat{\tilde{A}}' & = \sum_{k=1}^K s_{k} U^{(k)} \\ 
      & = \sum_{k=1}^K s_{k} ((1 - \alpha)^k (\hat{A}')^k +  \sum_{l=1}^{k} \alpha (1 - \alpha)^{l-1} (\hat{A}')^{l-1})
\end{split}
\end{myequation}
As a result, each coefficient $c_{k}$ becomes as follows:
\begin{myequation}
c_{k} = \left\{
\begin{array}{lcl}
\sum_{l=1}^{K} s_{l} \alpha  &      & { k = 0}\\
s_{k} (1 - \alpha)^k + \sum_{l=k+1}^{K} s_{l} \alpha (1 - \alpha)^{k}             &      & {1 \leq k \leq K }\\
\end{array}
\right.
\end{myequation}

Obviously, the coefficients $\{c_{k}\}$ are decided by both $\alpha$ and $\{s_{k}\}$.
Since $\alpha$ is a fixed hyper-parameter, we design $\{s_{k}\}$ as learnable values to make the coefficients $\{c_{k}\}$ of the polynomial optimizable.
We constrain that $\sum_{k=1}^{K} {s_{k}} = 1$ ($0 < s_{k} < 1$) and design a \textbf{coefficient-attention model} to learn $\{s_{k}\}$.
The coefficient-attention model introduces $K$ trainable scalar parameters $\{\hat{s}_{k}\}$ with the same initial value zero and applies the following operation on them:
\begin{myequation}\label{eq:softmax}
s_{k} = \frac{\exp(leaky\_relu(\hat{s}_{k}))}{\sum_{l=1}^{K} \exp(leaky\_relu(\hat{s}_{l}))}
\end{myequation}
%
Since $\{\hat{s}_{k}\}$ are initialized with the same value (zero), the above equation will assign the same value $\frac{1}{K}$ for all the $s_{k}$ before training.
Therefore, the initial values of the coefficients are decided by the Personalized PageRank algorithm.
During training, the $K$ parameters $\{\hat{s}_{k}\}$ are optimized, and accordingly, the coefficients $\{c_{k}\}$ are adaptively adjusted.

\subsubsection{Efficient Sparse Message Propagation for Extended Neighbors.} 
In order to learn low-dimensional semantic representations of vertices, we adopt the predict-then-propagate framework~\cite{DBLP:conf/iclr/KlicperaBG19}, which first projects input vertices into a low-dimensional semantic space with an encoder and then performs message passing to aggregate semantic information from vertices' context with the guidance of an extended propagation scheme.
In detail, our model first employs a multilayer perceptron (MLP) model $f_{mlp}$ to encode raw feature vectors of vertices into low-dimensional vertex-independent semantic feature vectors $H^{(0)} \in \mathbb{R}^{|V| \times d_{z}}$ ($d_{z} \ll |V|$) as follows:
\begin{myequation}
  H^{(0)} = f_{mlp}(X)
\end{myequation}
%
%
Then, our model aggregates the context information from the extended neighbors of vertices with the extended propagation polynomial $\hat{\tilde{A}}$ as follows:
\begin{myequation}\label{eq:dense_Z}
\begin{split}
Z & = \sigma(\hat{\tilde{A}} H^{(0)}) \\ 
  & = \sigma(\sum_{k=1}^K s_{k} ((1 - \alpha)^k \hat{A}^k +  \sum_{l=1}^{k} \alpha (1 - \alpha)^{l-1} \hat{A}^{l-1}) H^{(0)})
\end{split}
\end{myequation}
where $\sigma$ is the activation function.
In many real-world dataset, although $\hat{A}$ is usually a $|V| \times |V|$ sparse matrix, $\hat{A}^k (k \geq 2)$ and $\hat{\tilde{A}}$ may be a $|V| \times |V|$ dense matrix, which may not retain the sparsity of the computation.
For example, if $\hat{\tilde{A}}$ is dense, it requires large memory of size $|V| \times |V|$ to store the dense $\hat{\tilde{A}}$, and the time complexity of the dense matrix multiplication operation $\hat{\tilde{A}} H^{(0)}$ becomes $O(|V|^{2}d_{z})$.
%
%
%
To address the problem, our model computes $Z$ in an equivalent power iteration form as follows:
\begin{myequation}\label{eq:power_iter_h}
\begin{split}
H^{(k)} & = (U^{(k)})'H^{(0)} \\ 
        & = (1 - \alpha) \hat{A} (U^{(k-1)})'H^{(0)} + \alpha H^{(0)}\\
        & = (1 - \alpha) \hat{A} H^{(k-1)} + \alpha H^{(0)}
\end{split}
\end{myequation}
\begin{myequation}\label{eq:z_sum}
\begin{split}
Z & = \sigma(\hat{\tilde{A}} H^{(0)}) \\ 
  & = \sigma(\sum_{k=1}^K s_{k} (U^{(k)})' H^{(0)}) \\
  & = \sigma(\sum_{k=1}^K s_{k} H^{(k)}) 
\end{split}
\end{myequation}
In Equation~\ref{eq:power_iter_h}, $\hat{A}$ is a sparse matrix and $H^{(k)}$ is always a $|V| \times d_{k}$ dense matrix.
Therefore, the time complexity of $\hat{A} H^{(k-1)}$ is $O(|E|d_{z})$, where $|E|$ is the number of edges, and the time complexity for computing $Z$ is $O((|E| + |V|)d_{z})$.
In addition, the space complexity is reduced to $O(|E|d_{z} + |V|d_{z})$.
Obviously, the power iteration form can retain the sparsity of the graph and is therefore more efficient than the original form (Equation~\ref{eq:dense_Z}).
\subsection{Negative-free Entropy-aware Graph Contrastive Loss}

We propose a negative-free entropy-aware graph contrastive loss to explicitly take advantage of unlabeled data for semi-supervised learning.
Our contrastive loss constrains our GNN model to learn consistent semantics for the same vertex across multiple randomly augmented views of a graph. 
Our model adopts a simple yet effective dropout-based augmentation strategy, which perturbs the input graph to obtain augmented views by simply enabling the built-in feature-level dropout and edge-level dropout of the GNN model.

Given two different augmented views $\tilde{Z}^{[a]}$ and $\tilde{Z}^{[b]}$, we choose the negative cosine similarity to measure the semantic distance between vertices as follows:
\begin{myequation}
  D(\tilde{Z}_i^{[a]}, \tilde{Z}_i^{[b]}) = -\frac{\tilde{Z}_i^{[a]}}{\lVert \tilde{Z}_i^{[a]} \rVert_{2}} \cdot \frac{\tilde{Z}_i^{[b]}}{\lVert \tilde{Z}_i^{[b]} \rVert_{2}}
\end{myequation}
where $\tilde{Z}_i^{[a]}$ and $\tilde{Z}_i^{[b]}$ are the learned vertex representations of the $i_{th}$ vertex $v_i$ of the two different augmented views, and $\lVert \cdot \rVert_{2}$ denotes the L2-norm operation~\cite{DBLP:conf/cvpr/WuXYL18}.
For each training step, our framework obtains $M$ different augmented views and aims to minimize the following negative-free contrastive loss to preserve the semantic consistency of different augmented views:
\begin{myequation}
\small
\begin{split}
\mathcal{L}_{cl} & = \sum_{i=1}^{|V|} \frac{2}{|V| M^2} \sum_{a=1}^M \sum_{b=1}^M D(\tilde{Z}_i^{[a]}, stop\_grad(\tilde{Z}_i^{[b]})) \\
  & = \sum_{i=1}^{|V|} \frac{2}{|V| M^2} \sum_{a=1}^M \sum_{b=1}^M -\frac{\tilde{Z}_i^{[a]}}{\lVert \tilde{Z}_i^{[a]} \rVert_{2}} \cdot stop\_grad (\frac{\tilde{Z}_i^{[b]}}{\lVert \tilde{Z}_i^{[b]} \rVert_{2}})
\end{split}
\end{myequation}
where $stop\_grad$ operation forces the encoder for $\tilde{Z}_i^{[b]}$ to receive no gradient from $\tilde{Z}_i^{[b]}$.
As mentioned in \cite{DBLP:journals/corr/abs-2011-10566}, the introduction of $stop\_grad$ can prevent the optimizer from finding degenerated solutions caused by collapsing.
Note that $\mathcal{L}_{cl}$ is a negative-free contrastive loss, which does not require negative samples for contrastive learning.
Therefore, our model does not require extra memory to maintain a large number of negative samples, and does not rely on carefully designed negative sampling strategies.

For semi-supervised classification tasks, we extend the contrastive loss and design an entropy-aware contrastive loss to take into account a common assumption in many semi-supervised classification tasks that the model should make predictions with low entropy.
In order to introduce the low-entropy regularization, we leverage the sharpening technique~\cite{NEURIPS2019_1cd138d0,DBLP:conf/nips/FengZDHLXYK020} to extend $\mathcal{L}_{cl}$.
First, we introduce a temperature hyper-parameter $\tau$ ($0 < \tau \leq 1$) into Equation~\ref{eq:z_sum} to sharpen the predicted distribution and obtain predictions with lower entropy as follows:
\begin{myequation}
  \hat{Z} = \sigma(\frac{\sum_{k=1}^K s_{k} H^{(k)}}{\tau}) 
\end{myequation}
where we can set $\tau < 1$ to encourage the sharpened distribution $\hat{Z}$ to have lower entropy than $Z$.
%
%
We use $\hat{\tilde{Z}}^{[\cdot]}$ to denote the sharpened prediction of an augmented view.
Then, with the introduction of the sharpened predictions of augmented views, we extend the negative-free contrastive loss $\mathcal{L}_{cl}$ and design a negative-free entropy-aware contrastive loss $\mathcal{L}_{ecl}$ as follows:
\begin{myequation}
\mathcal{L}_{ecl} = \sum_{i=1}^{|V|} \frac{2}{|V| M^2} \sum_{a=1}^M \sum_{b=1}^M -\frac{\tilde{Z}_i^{[a]}}{\lVert \tilde{Z}_i^{[a]} \rVert_{2}} \cdot stop\_grad (\frac{\hat{\tilde{Z}}_i^{[b]}}{\lVert \hat{\tilde{Z}}_i^{[b]} \rVert_{2}})
\end{myequation}
%
%
Note that $\mathcal{L}_{cl}$ is a special case of $\mathcal{L}_{ecl}$ when $\tau = 1$.
We find that the low-entropy assumption may fail on some datasets, and we can set $\tau = 1$ to deal with the problem.

\subsection{Model Optimization}

We combined the supervised loss function $\mathcal{L}_{sup}$ with $\mathcal{L}_{ecl}$ and the L2 loss $\mathcal{L}_{L2}$ as $\mathcal{L}$ as follows:
\begin{myequation}
  \mathcal{L} = \mathcal{L}_{sup} + \psi_{ecl} \mathcal{L}_{ecl} + \psi_{L2} \mathcal{L}_{L2}
\end{myequation}
where we use the cross-entropy loss as $\mathcal{L}_{sup}$ for vertex classification tasks, and $\psi_{ecl}$ and $\psi_{L2}$ are the weights for $\mathcal{L}_{ecl}$ and $\mathcal{L}_{L2}$, respectively.
The L2 loss is employed to avoid overfitting.
We employ the Adam optimizer~\cite{DBLP:journals/corr/KingmaB14} to minimize $\mathcal{L}$ to jointly optimize $\mathcal{L}_{sup}$, $\mathcal{L}_{ecl}$, and $\mathcal{L}_{L2}$.
More details about the training can be found in Appendix~\ref{appendix:implement}.


\subsection{Relation of CAPGNN to APPNP}\label{sec:capgnn_and_appnp}

As with CAPGNN, the propagation scheme of APPNP can also be modeled as a polynomial of $\hat{A}'$:
\begin{myequation}
\begin{split}
\hat{\tilde{A}}'_{APPNP} & = U^{(K)} \\ 
& = (1 - \alpha)^K (\hat{A}')^K +  \sum_{l=1}^{K} \alpha (1 - \alpha)^{l-1} (\hat{A}')^{l-1}
\end{split}
\end{myequation}
Power operation does not change the eigenvectors of $\hat{A}'$, since $(\hat{A}')^k \vec{r} = \lambda^k \vec{r}$, where $\vec{r}$ is an eigenvector of $\hat{A}'$ and $\lambda$ is the corresponding eigenvalue.
Therefore, the propagation scheme of APPNP should be equivalent to that of CAPGNN if they share the same eigenvalue set.
Given $\vec{r}$ and $\lambda$, the corresponding eigenvalue of $\hat{\tilde{A}}'_{APPNP}$ can be obtained as follows when $K \to + \infty$:
\begin{myequation}
\begin{split}
\lim_{K \to + \infty} \lambda_{APPNP} & = (1 - \alpha)^K (\lambda)^K +  \sum_{l=1}^{K} \alpha (1 - \alpha)^{l-1} (\lambda)^{l-1} \\
& = \frac{\alpha}{1 - (1 - \alpha)\lambda}
\end{split}
\end{myequation}
Similarly, we can infer the corresponding eigenvalue for the propagation scheme of CAPGNN $\hat{\tilde{A}}'$ as follows (based on Equation.~\ref{eq:combined_A}):
\begin{myequation}
\begin{split}
\lambda_{CAPGNN} & = \sum_{k=1}^K s_{k} ((1 - \alpha)^k (\lambda)^k +  \sum_{l=1}^{k} \alpha (1 - \alpha)^{l-1} (\lambda)^{l-1})
\end{split}
\end{myequation}
%
%
If we remove the coefficient-attention model of CAPGNN and use static coefficient $\frac{1}{K}$ for $\{s_{k}\}$ instead, we denote the model as CPGNN, and its eigenvalue $\lambda_{CPGNN}$ becomes as follows when $K \to + \infty$:
\begin{myequation}
\begin{split}
\lim_{K \to + \infty} \lambda_{CPGNN} & = \sum_{k=1}^K \frac{1}{K} ((1 - \alpha)^k (\lambda)^k +  \sum_{l=1}^{k} \alpha (1 - \alpha)^{l-1} (\lambda)^{l-1}) \\
& = \frac{\alpha}{1 - (1 - \alpha)\lambda} \\
& = \lim_{K \to + \infty} \lambda_{APPNP}
\end{split}
\end{myequation}
which verifies that CAPGNN's propagation scheme is equivalent to APPNP's if we disable the coefficient-attention model.
As mentioned in Section~\ref{sec:detail_adapt_prop}, the coefficient-attention model initializes $s_{k}$ with the same value $\frac{1}{K}$.
This results in that our model will use APPNP's propagation scheme at the beginning of training.
Therefore, our model can benefit from the superior initial coefficients provided by Personalized PageRank.

\begin{table*}[tb]
  \centering
  \caption{Statistics of Datasets.}
  \begin{tabular}{|l|c|c|c|c|c|c|c|c|c|}\hline
    Dataset          & Vertices & Edges  & Features & Classes \\\hline
    Cora             & 2708     & 5278   & 1433     & 7       \\\hline
    Citeseer         & 3327     & 4552   & 3703     & 6       \\\hline
    Pubmed           & 19717    & 44324  & 500      & 3       \\\hline
    Amazon-Computers & 13752    & 245861 & 767      & 10      \\\hline
    Amazon-Photo     & 7650     & 119081 & 745      & 8       \\\hline
    \end{tabular}
  \label{tab:datasets_statistics}
\end{table*}

\begin{table*}[t]
  \caption{Summary of Results in Terms of Classification Accuracies.}
  \label{tab:performance}
  \centering
  \begin{tabular}{|l|c|c|c|c|c|c|c|c|c|c|}\hline
\multirow{2}{*}{\diagbox[width=10.0em]{Dataset}{Method}}    &\multicolumn{8}{c|}{Baseline} & \multicolumn{2}{c|}{CAPGNN}\\\cline{2-11}

                 & GCN   & GAT   & APPNP & S$^2$GC & ADSF-RWR & GraphMix & CAD-Net & GRAND           & CAPGCN         & CAPGAT         \\\hline
Cora             & 0.815 & 0.830 & 0.836 & 0.835   & 0.854    & 0.839    & 0.843   & 0.854           & 0.861          & \textbf{0.862} \\\hline
Citeseer         & 0.703 & 0.725 & 0.722 & 0.736   & 0.740    & 0.747    & 0.741   & 0.754           & 0.751          & \textbf{0.758} \\\hline
Pubmed           & 0.790 & 0.790 & 0.797 & 0.802   & 0.812    & 0.811    & 0.823   & \textbf{0.827}  & \textbf{0.827} & \textbf{0.827} \\\hline
Amazon-Computers & 0.813 & 0.814 & 0.818 & 0.820   & --       & 0.818    & 0.821   & 0.823           & 0.827          & \textbf{0.834} \\\hline
Amazon-Photo     & 0.901 & 0.898 & 0.906 & 0.902   & --       & 0.901    & 0.909   & 0.911           & 0.915          & \textbf{0.920} \\\hline

    



  \end{tabular}
\end{table*}

\section{Experiments}
In this section, we conduct experiments on benchmark datasets to verify the effectiveness of CAPGNN.

\subsection{Datasets and Baselines.}

We consider a task of semi-supervised node classification on 5 benchmark datasets, including 3 widely used citation networks (Cora, CiteSeer, and Pubmed) and 2 recommendation networks (Amazon-Computers and Amazon-Photo).
For the citation networks, the vertices, edges, vertex features, and vertex labels correspond to documents, citation links, the text of documents in the form of bag-of-words, and fields of study of documents, respectively.
For the recommendation networks, the vertices represent products, the edges represent the relation that two products are frequently bought together, the vertex features are the product reviews in the form of bag-of-words, and the vertex labels are product categories.
Table~\ref{tab:datasets_statistics} summarizes the statistics of the 5 datasets.


For the citation networks, we follow \cite{DBLP:conf/iclr/KipfW17} and employ the predefined data split setting.
For the recommendation networks, we use the same data split setting as in \cite{DBLP:conf/aaai/LimUCJC21}, where for each class, 20 and 30 vertices are randomly selected for train and validation, respectively, and remaining vertices are used for test.
Both split settings split each dataset into a train, valid, and test set, where the valid set is only used for early stopping.
%

Our baselines include GCN~\cite{DBLP:conf/iclr/KipfW17}, GAT~\cite{DBLP:conf/iclr/VelickovicCCRLB18}, APPNP~\cite{DBLP:conf/iclr/KlicperaBG19}, S$^2$GC~\cite{DBLP:conf/iclr/ZhuK21}, ADSF-RWR~\cite{DBLP:conf/iclr/0001ZWZ20}, GraphMix~\cite{DBLP:conf/aaai/VermaQKLBKT21}, CAD-Net~\cite{DBLP:conf/aaai/LimUCJC21}, and GRAND~\cite{DBLP:conf/nips/FengZDHLXYK020}.
Among them, CAD-Net and GRAND are state-of-the-art baselines that can take advantage of both extended neighbor information and unlabeled data.
For the citation datasets, we adopt the performance of baselines reported in the original paper.
For the recommendation datasets, due to the lack of unified split setting, we evaluate the baselines using the official implementations with carefully tuned hyper-parameters.
Due to some technical problems, we are not able to run all the official implementations and therefore some of the results on the recommendation datasets are missing.
In terms of the parameter settings, as with most GNN research, we do not compare different approaches with unified parameter settings.
The reason is that different baselines usually rely on different parameter settings to achieve their best performance due to their different mechanisms.
The detailed parameter setting of our model is in Appendix~\ref{appendix:params} and source code~\footnote{https://github.com/hujunxianligong/CAPGNN}.
In addition, we run our model and each variant 100 times and report the mean accuracy scores.




%


\subsection{Performance Analysis} 
Based on the results shown in Table~\ref{tab:performance}, we have the following observations:
(1) GCN and GAT, which focus on local propagation schemes and aggregate neighbor information within 2-hops, perform worse than most other baselines, which rely on extended propagation schemes to exploit neighbors within K hops ($K > 2$), showing the necessity of extended propagation schemes.
(2) Although GraphMix does not involve extended propagation schemes, it achieves competitive performance with most other baselines by introducing regularization for unlabeled data, demonstrating that the explicit utilization of unlabeled data is also crucial for building effective GNNs.
(3) CAD-Net, GRAND, and our models show superior performance over other baselines.
They not only utilize extended propagation schemes, but also explicitly exploit unlabeled data for training, where CAD-Net makes use of the pseudo labels of unlabeled data, and GRAND and our models apply regularizations on unlabeled data.
This shows that GNNs with extended propagation schemes can further benefit from the explicit utilization of unlabeled data.
(4) Our models beat all the baselines, showing the superiority of our adaptive extended propagation scheme and the contrastive loss.
(5) CAPGAT performs slightly better than CAPGCN, showing that our extended propagation scheme can benefit from the dynamic learning of local context distribution.

\begin{figure}[!bp]
  \centering
  \includegraphics[scale=0.31]{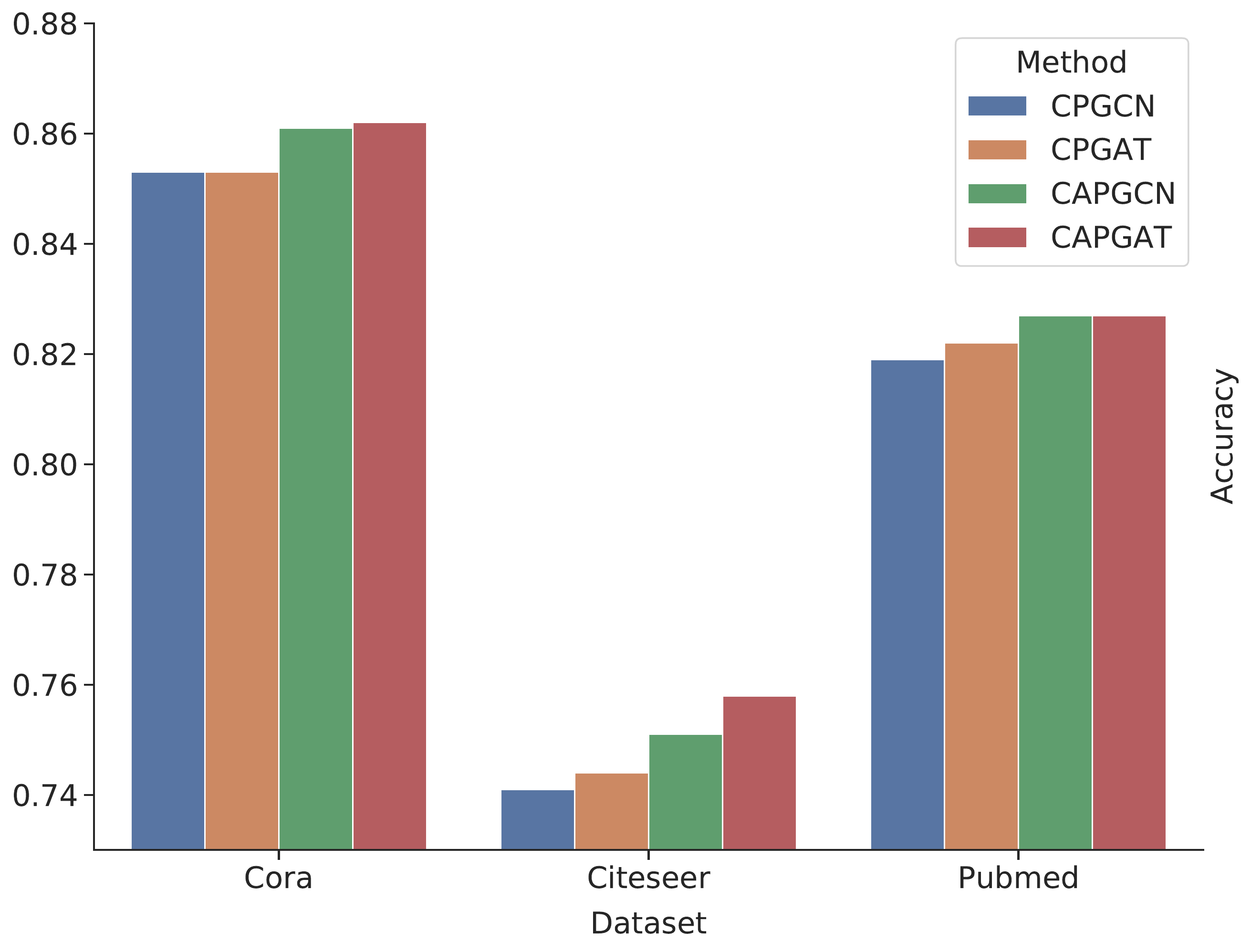}
  \caption{Impact of Adaptive Extended Propagation Scheme.}
  \label{fig:ablation_adaptive_prop}
  \end{figure}

\begin{figure}[!tbp]
  \centering
  \subfigure[CAPGCN]{
  \label{xxxxx} 
  \includegraphics[width=2.9in]{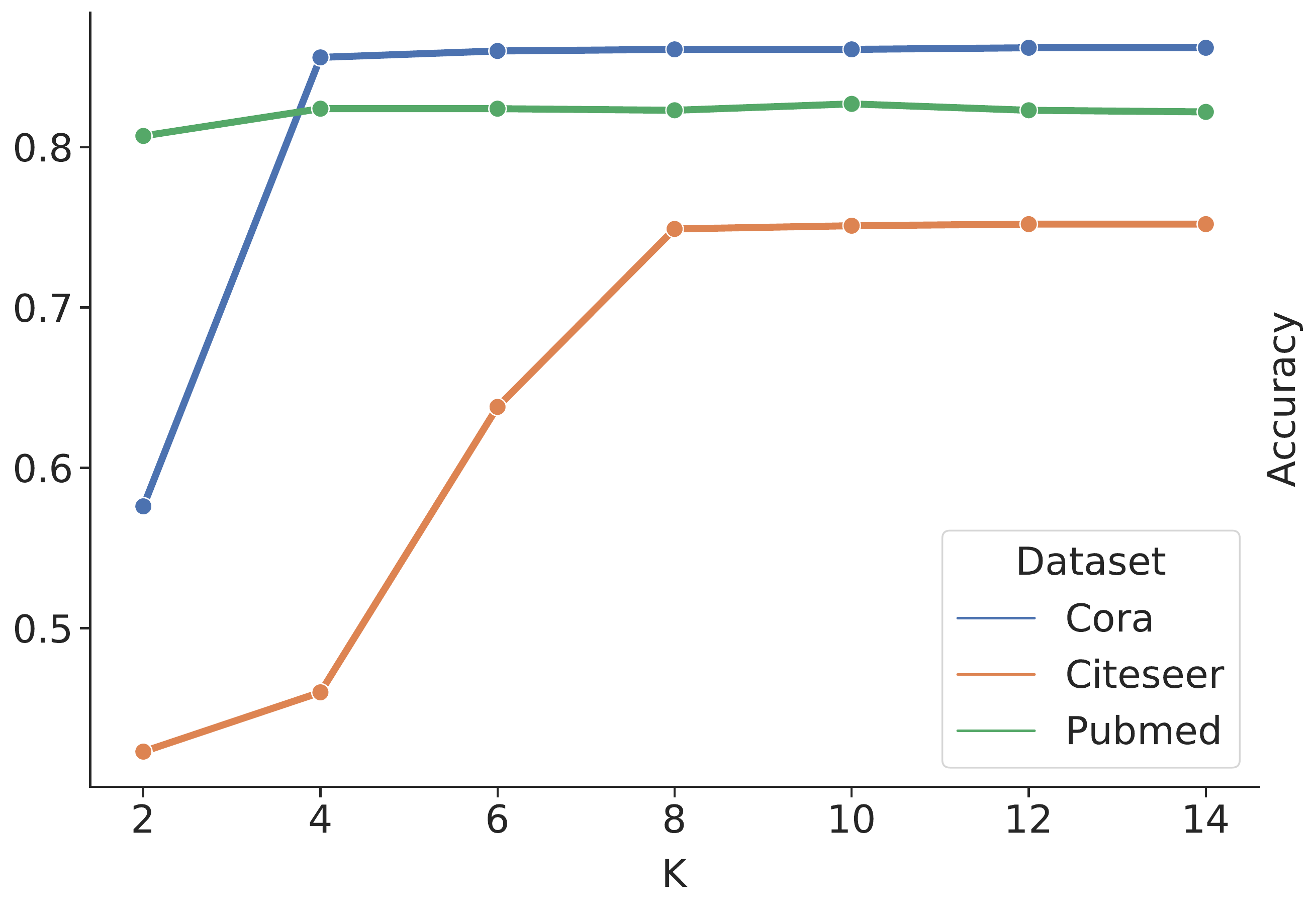}
  }
  \subfigure[CAPGAT]{
  \label{xxxxx} 
  \includegraphics[width=2.9in]{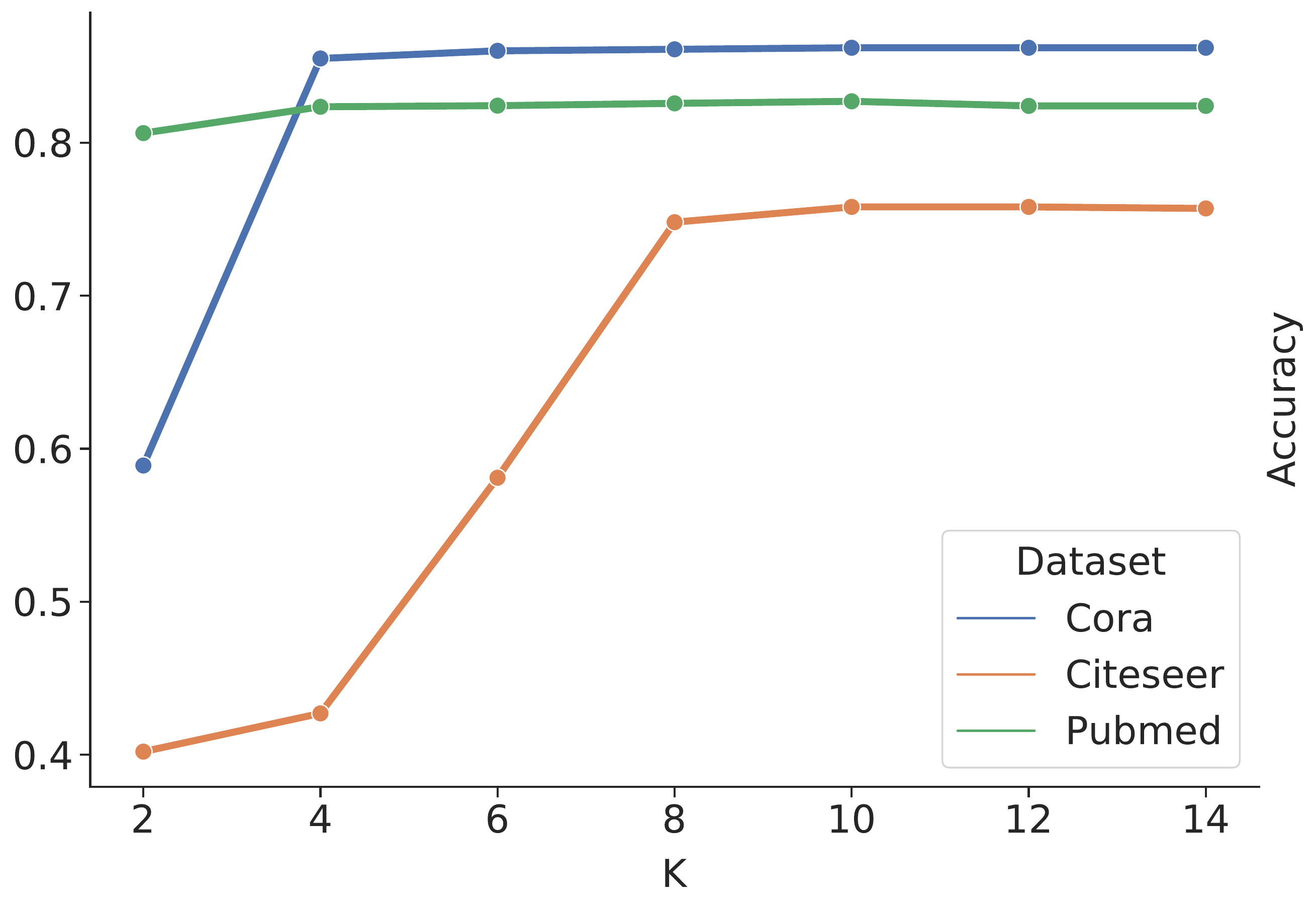}
  }
  %
  \caption{Impact of Number of Iterations $K$.}
  \label{fig:impact_num_layers} 
  \end{figure}

\subsection{Detailed Analysis}

\subsubsection{Impact of Adaptive Extended Propagation Scheme.}
To verify the effectiveness of our adaptive extended propagation scheme, we build a variant model named CPGNN (including CPGCN and CPGAT) by replacing our propagation scheme with APPNP's propagation scheme, namely $U^{(K)}$ mentioned in Section~\ref{sec:capgnn_and_appnp}.
As mentioned in Section~\ref{sec:capgnn_and_appnp}, a CPGNN model is equivalent to a degraded CAPGNN model, which disables its adaptive propagation ability by removing its coefficient-attention model and using static coefficient $\frac{1}{K}$ for $\{s_{k}\}$ instead.
The performance of CPGNN and CAPGNN is shown in Figure~\ref{fig:ablation_adaptive_prop}.
The results show that CAPGNN can consistently outperform CPGNN, showing the effectiveness of our adaptive extended propagation scheme.

\subsubsection{Impact of Number of Iterations.}
We vary the number of iterations $K$ from 2 to 14 and report the performance in Figure~\ref{fig:impact_num_layers}.
In the beginning, CAPGNN obtains dramatic performance improvements with the increase of $K$, and the performance can still be improved gradually with more iterations, verifying that our model can effectively exploit high-order neighbor information.
In addition, the performance of CAPGNN remains steady when $K \geq 10$, showing that CAPGNN is not sensitive to $K$ when sufficient iterations are provided.
Therefore, we can simply set $K$ to 10 on all the datasets and do not need to choose a $K$ carefully for each dataset.

\begin{figure}[!tb]
  \centering
  \subfigure[Cora]{
  \label{xxxxx} 
  \includegraphics[width=3.1in]{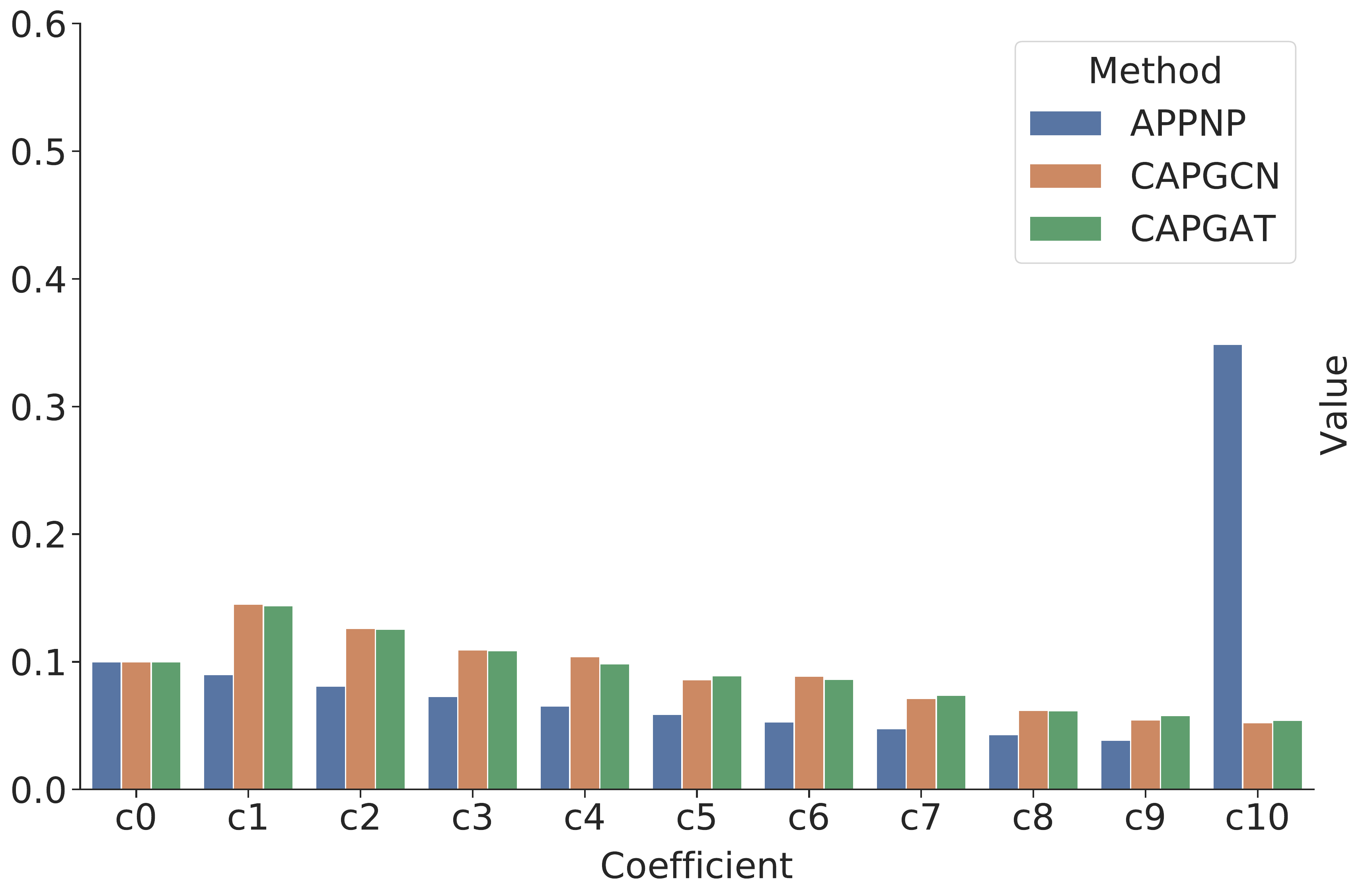}
  }
  %
  \subfigure[Citeseer]{
  \label{xxxxx} 
  \includegraphics[width=3.1in]{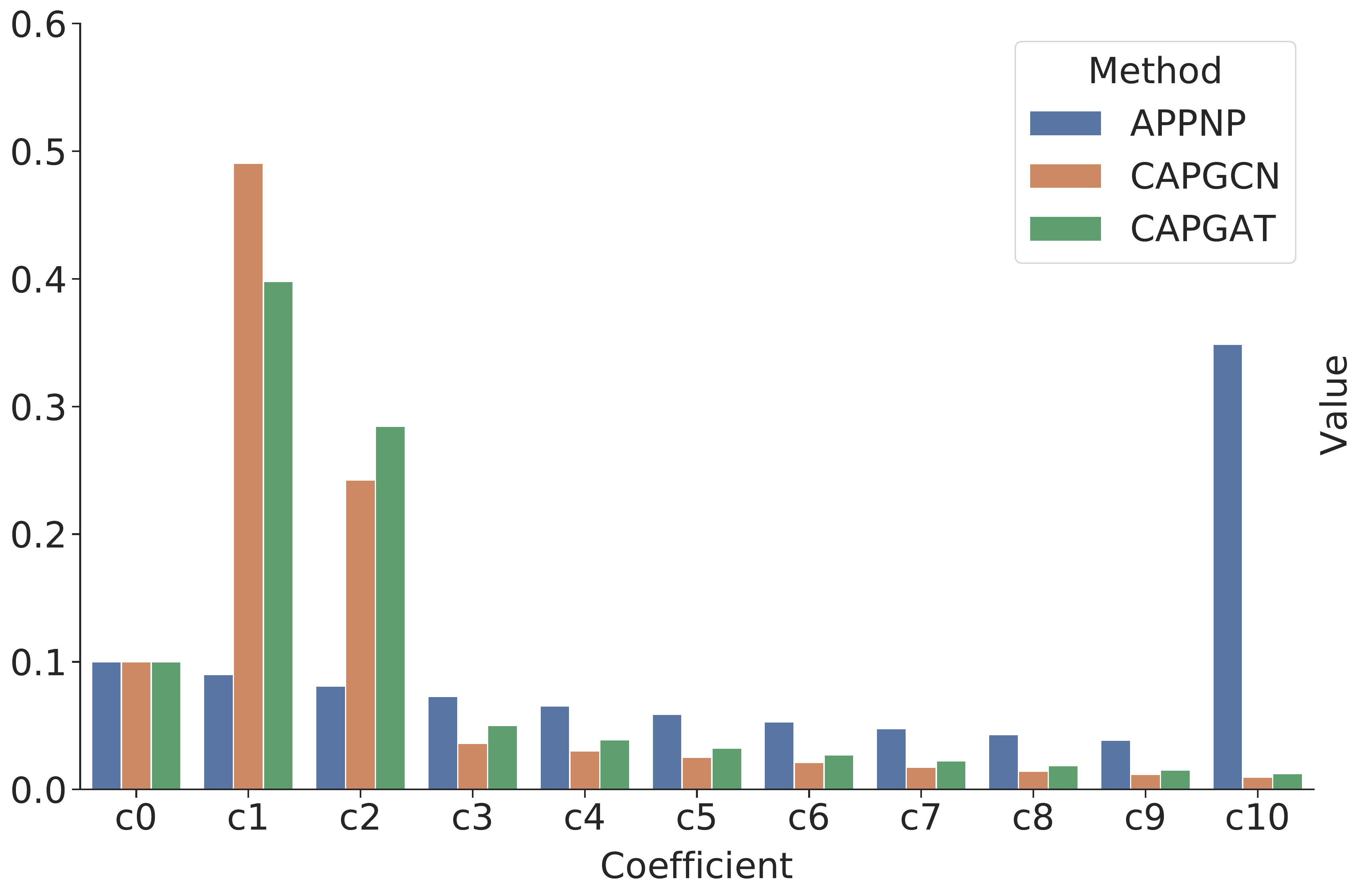}
  }
  \subfigure[Pubmed]{
  \label{xxxxx} 
  \includegraphics[width=3.1in]{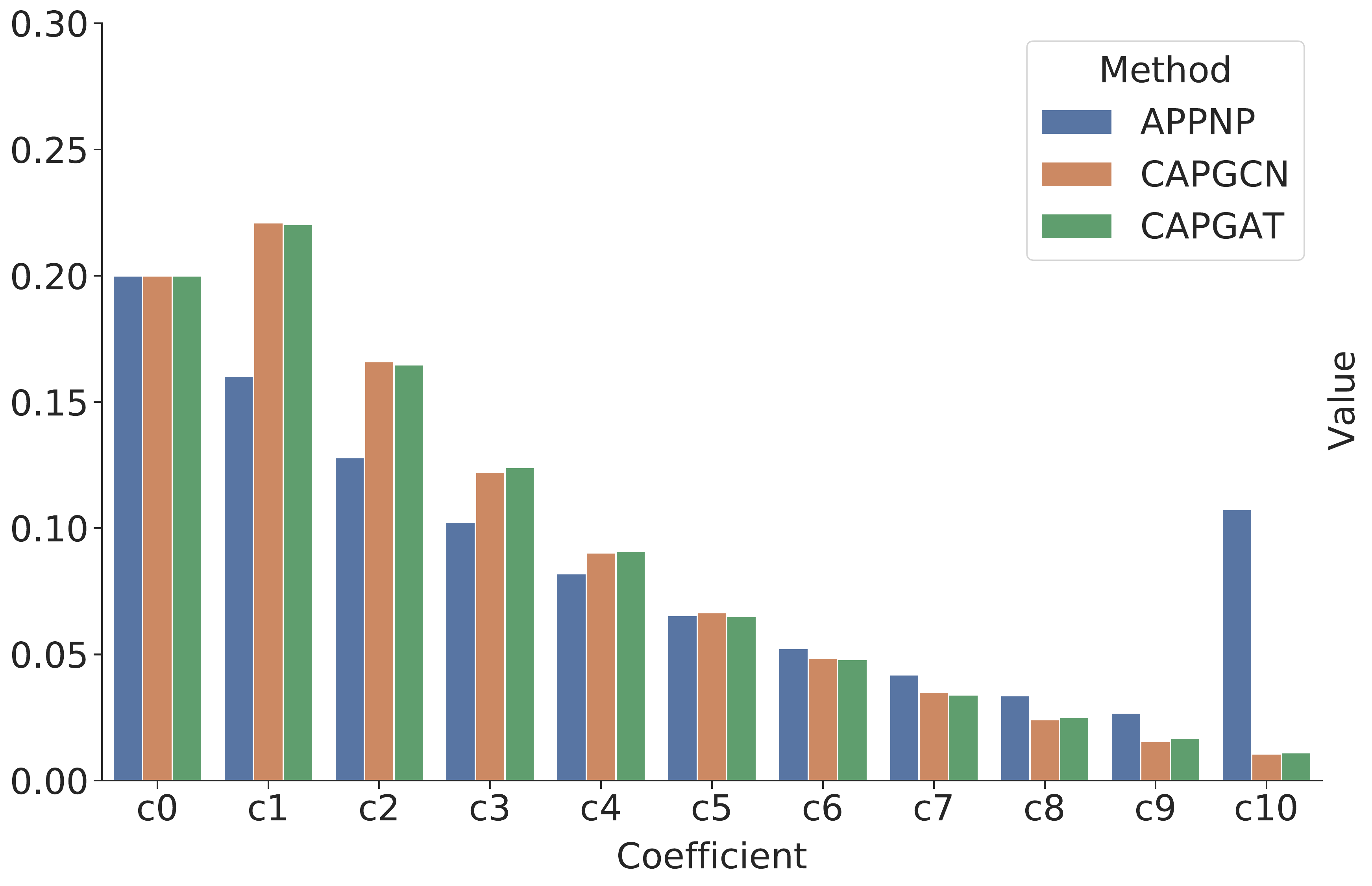}
  }
  %
  \caption{Visualization of Coefficients of Propagation Polynomial.}
  \label{fig:visual_coefs} 
\end{figure}

\subsubsection{Visualization of Coefficients of Propagation Polynomial.}
We visualize the coefficients $\{c_{k}\}$ of the propagation polynomial of APPNP and CAPGNN in Figure~\ref{fig:visual_coefs}.
Since we set $\alpha$ to 0.1 on Cora and Citeseer and 0.2 on Pubmed, the distribution of coefficients on Pubmed is quite different from that on Cora and Citeseer.
Obviously, there are significant differences between the distribution of coefficients of APPNP and CAPGNN.
The most notable difference is that APPNP usually assigns a large value for $c_{10}$, while CAPGNN's $c_k$ usually gradually declines when increasing $k$ from 2 to 10.
This shows that APPNP highly relies on the $K-$hop neighbors, while CAPGNN relies more on the combination of different hops of neighbors.
This may be one of the reasons why CAPGNN's propagation scheme shows superior performance over APPNP's.

\begin{figure}[tb]
  \centering
  \includegraphics[scale=0.32]{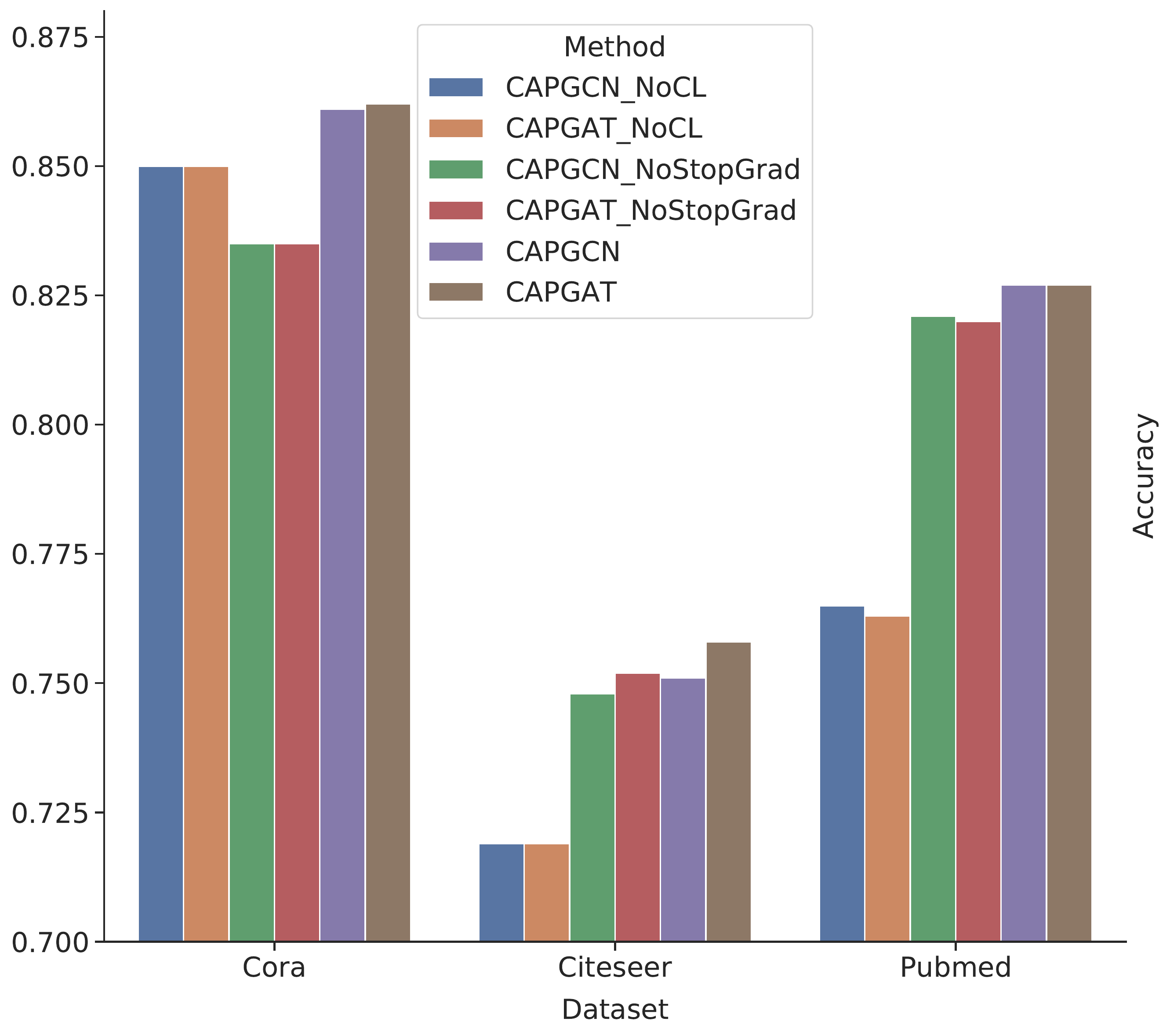}
  \caption{Impact of Negative-free Entropy-aware Contrastive Loss.}
  \label{fig:ablation_all_cl}
  \end{figure}

  \begin{figure}[tb]
    \centering
    \subfigure[Cora]{
    \label{xxxxx} 
    \includegraphics[width=2.9in]{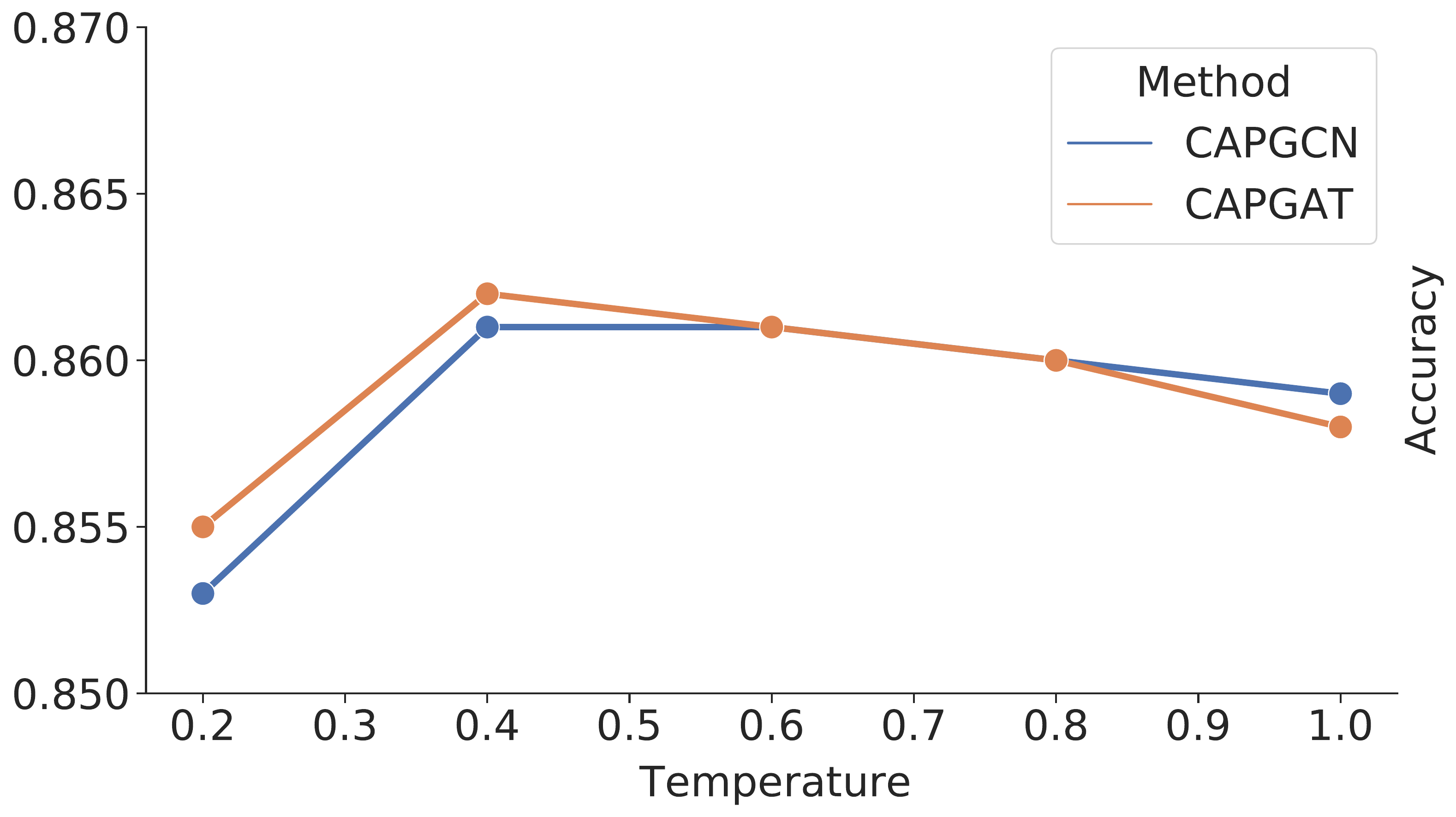}
    }
    \subfigure[Citeseer]{
    \label{xxxxx} 
    \includegraphics[width=2.9in]{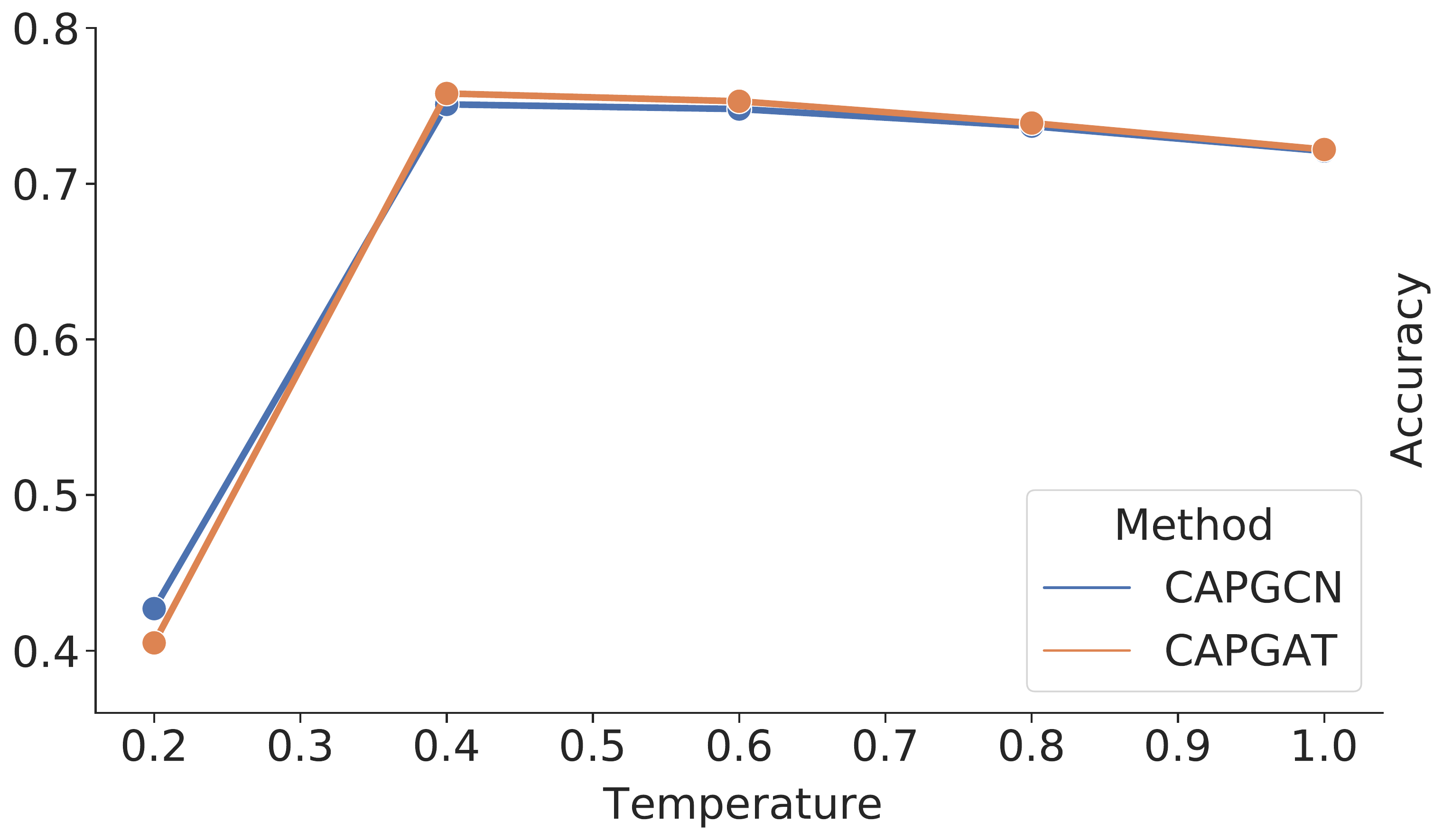}
    }
    \subfigure[Pubmed]{
    \label{xxxxx} 
    \includegraphics[width=2.9in]{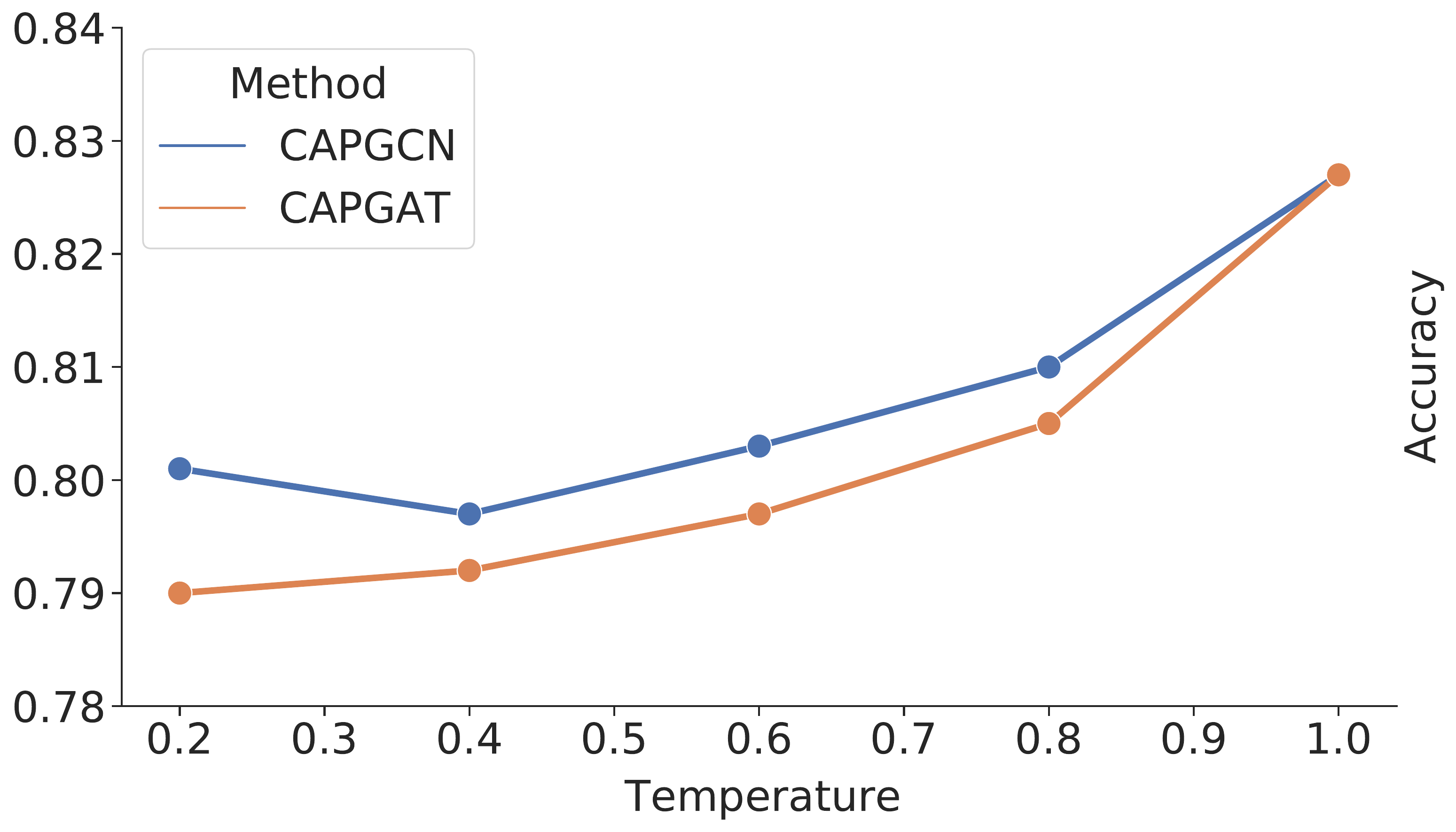}
    }
    %
    \caption{Impact of Low-entropy Assumption.}
    \label{fig:impact_low_entropy} 
    \end{figure}

\subsubsection{Impact of Negative-free Entropy-aware Contrastive Loss.}
To verify the effectiveness of the negative-free entropy-aware contrastive loss, we build a variant model GAPGNN\_NoCL, which removes the contrastive loss for training.
The performance of CAPGNN\_NoCL is shown in Figure~\ref{fig:ablation_all_cl}.
With the introduction of the contrastive loss, dramatic performance improvements can be observed in Figure~\ref{fig:ablation_all_cl}, which verifies the effectiveness of the contrastive loss.
In addition, to demonstrate the effectiveness of the $stop\_grad$ operation in the contrastive loss, we remove it from CAPGNN to build the variant model CAPGNN\_NoStopGrad.
%
%
We compare CAPGNN\_NoStopGrad and CAPGNN in Figure~\ref{fig:ablation_all_cl}.
The results show that CAPGNN\_NoStopGrad consistently performs worse than CAPGNN, showing the effectiveness of the $stop\_grad$ for optimization.
Although CAPGNN\_NoStopGrad obtains degenerated performance, it does not suffer from severe collapsing problems as mentioned in ~\cite{DBLP:journals/corr/abs-2011-10566}.
This may be because CAPGNN\_NoStopGrad can benefit from the supervised loss, which is not available for the self-supervised tasks in ~\cite{DBLP:journals/corr/abs-2011-10566}.

\subsubsection{Impact of Low-entropy Assumption.}
We also conduct experiments to verify the effectiveness of the low-entropy assumption in the contrastive loss.
We vary the temperature hyper-parameter $\tau$ from 0.2 to 1.0 to gradually reduce the requirements for low-entropy outputs, where the low-entropy assumption is totally removed when setting $\tau=1.0$.
The performance is reported in Figure~\ref{fig:impact_low_entropy}.
On the Cora and Citeseer datasets, our model achieves the best performance when $\tau=0.4$.
When $\tau$ varies from 0.4 to 1.0, the performance gradually decreases, showing that the contrastive loss can benefit from the low-entropy assumption on the two datasets.
When we set $\tau=0.2$ to apply a stronger low-entropy constraint, an obvious performance decline is observed, showing the necessity to choose a proper $\tau$ value.
On the Pubmed dataset, our model achieves the best performance when the low-entropy assumption is totally removed ($\tau=1.0$), showing that the low-entropy assumption may fail on some datasets.
As shown in Figure~\ref{fig:ablation_all_cl}, despite the failure of the low-entropy, our model still can benefit from the semantic consistency constraint of the contrastive loss and achieve better performance than the variants without the contrastive loss.



\section{Conclusions}

We propose an efficient yet effective end-to-end framework, namely Contrastive Adaptive Propagation Graph Neural Networks (CAPGNN).
Our model shows that GNNs can benefit from learnable extended propagation schemes that can adaptively adjust the influence of local and high-order neighbors.
In addition, we design a negative-free entropy-aware contrastive loss to explicitly take advantage of unlabeled data for training and achieve better performance, demonstrating that the explicit utilization of unlabeled data is likely to allow us to optimize extended propagation schemes with limited label information.
In the future, we plan to apply our approach to more real-world applications such as recommendation systems, social network analysis, and social media understanding.






\bibliographystyle{ACM-Reference-Format}
\bibliography{CAPGNN}



\appendix
\clearpage
\section{Implementation Details Appendix}\label{appendix:implement}
We list the pseudo-code of the forward process and training process of CAPGNN in Algorithm 1 and Algorithm 2, respectively.
Note that during training (Algorithm 2), we perform the \textbf{forward} method described in Algorithm 1 $M$ times with dropout enabled to obtain $M$ different augmented views for the negative-free contrastive loss.
After training, we should disable the dropout of the \textbf{forward} method for inference.
Besides, we can further speed up Algorithm 2 by executing the $M$ iterations of $forward$ in parallel with techniques such as graph mini-batch
, which are well supported by many GNN libraries such as tf\_geometric~\cite{DBLP:conf/mm/HuQFWZZX21} and torch\_geometric~\cite{DBLP:journals/corr/abs-1903-02428}.

\vspace{-1mm}
\begin{algorithm}[htb]
  \caption{Pseudocode of Forward of CAPGNN.}
  \label{alg:forward}
  \definecolor{codeblue}{rgb}{0.25,0.5,0.5}
  \lstset{
    backgroundcolor=\color{white},
    basicstyle=\fontsize{7.2pt}{7.2pt}\ttfamily\selectfont,
    columns=fullflexible,
    breaklines=true,
    captionpos=b,
    commentstyle=\fontsize{7.2pt}{7.2pt}\color{codeblue},
    keywordstyle=\fontsize{7.2pt}{7.2pt},
  }
  \begin{lstlisting}[language=python]
  # method: "CAPGCN" or "CAPGAT"
  # x: vertex feature matrix of input graph 
  # A: sparse adjacency matrix of input graph
  # dr_input, dr_mlp, dr_edge, dr_coef_att: dropout rates
  # affinity: sparse local affinity matrix

  # initialize coefficient-attention model
  coef_att_logits = Variable(zeros([K]))
  
  # forward with CAPGNN
  def forward():
      # applying dropout on input features
      x_ = Dropout(x, rate=dr_input)  
      
      A_ = A + I # add self-loop
      D_ = matrix_diag(A.sum(axis=-1) + 1) # degree matrix
      # @ denotes matrix multiplication
      gcn_affinity = pow(D_, -0.5) @ A_ @ pow(D_, -0.5)
  
      # CAPGCN and CAPGAT use different affinity matrix
      if method == "CAPGCN":
          affinity = gcn_affinity
      else:
          # compute the attention matrix of GAT
          T = gat_compute_attention(x_, A)
          # Renormalize T and combine it with gcn_affinity
          affinity = beta * gcn_affinity + \
                (1 - beta) * pow(D_, 0.5) @ T @ pow(D_, -0.5)
    
      # dropout is applied to the hidden layers of MLP
      h0 = MLP(x_, drop_rate=dr_mlp) # H^{(0)}
      h = h0 
      h_list = [] # {H^{k}|1 <= k <=K}
  
      # propagate with K iterations
      for _ in range(K):
          # apply dropout on sparse local affinity matrix
          affinity_ = Dropout(affinity, dr_edge)
          # Personalized PageRank-style propagation
          h = (1 - alpha) * affinity_ @ h  + alpha * h0
          # save the result of current iteration
          h_list.append(h) 
  
      # activate and normalize coefficient-attention scores
      # then, broadcast it from [K] to [K, |V|, d_z]
      coef_att_scores = softmax_and_broadcast(leaky_relu(coef_att_logits, negative_slope=0.2))
  
      # apply dropout on attention scores
      coef_att_scores = Dropout(coef_att_scores, rate=dr_coef_att)
  
      # apply attention scores on propagation results
      h_matrix = stack(h_list, axis=0) * coef_att_scores
      logits = h_matrix.sum(axis=0) # weighted sum
      return logits

  \end{lstlisting}
  \end{algorithm}

\begin{algorithm}[htb]
  \caption{Pseudocode of Training of CAPGNN.}
  \label{alg:train}
  \definecolor{codeblue}{rgb}{0.25,0.5,0.5}
  \lstset{
    backgroundcolor=\color{white},
    basicstyle=\fontsize{7.2pt}{7.2pt}\ttfamily\selectfont,
    columns=fullflexible,
    breaklines=true,
    captionpos=b,
    commentstyle=\fontsize{7.2pt}{7.2pt}\color{codeblue},
    keywordstyle=\fontsize{7.2pt}{7.2pt},
  }
  \begin{lstlisting}[language=python]
  # y: label of vertices
  # train_mask: boolean mask (True for training vertices)
  # t: temperature
  # ecl_weight, l2_weight: weights of losses

  # train CAPGNN
  def train_step(adam_optimizer):
      logits_list = [] # store outputs of M augmented views
      for _ in range(M): # M different augmented views
          # enable Dropout in forward for augmentation   
          logits = forward() 
          logits_list.append(logits)

      # compute supervised losses of M augmented views
      supervised_losses = [CrossEntropy(logits, y, train_mask) for logits in logits_list]
      
      # [|V|, M, d_z]
      logits_matrix = stack(logits_list, axis=1) 
      Z = softmax(logits_matrix, axis=-1)
      Z_ = softmax(logits_matrix / t, axis=-1) # temperature
      # [|V|, M, d_z] @ [|V|, d_z, M] -> [|V|, M, M]
      # l2_norm is used for cosine similarity
      ecl_losses = -2 * l2_norm(Z) @ stop_grad(transpose(l2_norm(Z_), [0, 2, 1]))
      
      l2_loss = compute_l2_loss() # exclude bias variables
      
      loss = mean(supervised_losses) + ecl_weight * mean(ecl_losses) + l2_weight * l2_loss
      adam_optimizer.minimize(loss) # optimize
  
  \end{lstlisting}
  \end{algorithm}

\section{Parameter Setting Appendix}\label{appendix:params}

In the experiments, we adopt the same CAPGNN architecture on all the datasets.
The architecture employs a two-layer MLP, where the dimensionalities of the output of the two layers are 64 and the number of classes, respectively.
The first layer of the MLP activates the output with ReLU activation function, and the second layer of the MLP does not activate the output.
Note that on the Pubmed dataset, we follow \cite{DBLP:conf/nips/FengZDHLXYK020} and apply Batch Normalization~\cite{DBLP:conf/icml/IoffeS15} on the input vertex features and the output of the first layer of the MLP.
For the affinity matrix of CAPGAT, we set the trade-off parameter $\beta$ to 0.3.
For the adaptive extended propagation scheme, we set the number of iterations $K$ to 10 and the negative slope coefficient of leaky\_relu in Equation~\ref{eq:softmax} to 0.2 for all the datasets.
In terms of the contrastive loss, we set the number of augmented views $M$ to 8, and set the weight of the contrastive loss $\psi_{ecl}$ to 1.0.
For training, we train our model with no more than 2000 epochs on each dataset and adopt the early stopping strategy~\cite{DBLP:conf/iclr/VelickovicCCRLB18} with a patience of 200 epochs.
%
%
The detailed settings of hyper-parameters are listed as follows.
Note that $lr$ denotes the learning rate, and the hyper-parameters $dr\_input$, $dr\_mlp$, $dr\_edge$, and $dr\_coef\_att$ are dropout rates mentioned in Algorithm 1:
\vspace{-1mm}
\begin{itemize}
\small
\item \textbf{Cora:} $lr=1e-2$, $\psi_{L2}=1e-3$, $\alpha=0.1$, $dr\_input=0.8$, $dr\_mlp=0.9$, $dr\_edge=0.7$, $dr\_coef\_att=0.3$, $\tau=0.4$.
\item \textbf{Citeseer:} $lr=1e-2$, $\psi_{L2}=1e-3$, $\alpha=0.1$, $dr\_input=0.5$, $dr\_mlp=0.1$, $dr\_edge=0.0$, $dr\_coef\_att=0.3$, $\tau=0.4$.
\item \textbf{Pubmed:} $lr=2e-1$, $\psi_{L2}=2e-3$, $\alpha=0.2$, $dr\_input=0.1$, $dr\_mlp=0.15$, $dr\_edge=0.1$, $dr\_coef\_att=0.3$, $\tau=1.0$.
\item \textbf{Amazon-Computers:} $lr=1e-2$, $\psi_{L2}=1e-3$, $\alpha=0.1$, $dr\_input=0.4$, $dr\_mlp=0.6$, $dr\_edge=0.1$, $dr\_coef\_att=0.3$, $\tau=0.4$.
\item \textbf{Amazon-Photo:} $lr=1e-2$, $\psi_{L2}=1e-3$, $\alpha=0.1$, $dr\_input=0.6$, $dr\_mlp=0.7$, $dr\_edge=0.0$, $dr\_coef\_att=0.3$, $\tau=0.4$.
\end{itemize}


\end{document}